\documentclass{article}
\usepackage{spconf,amsmath,graphicx}
\usepackage{caption}
\usepackage{graphicx}
\usepackage{array}
\usepackage{amsmath}
\usepackage{nicefrac}
\usepackage{amsfonts}
\usepackage{amssymb}
\usepackage{amsthm}
\usepackage{algorithm}
\usepackage[noend]{algpseudocode}
\usepackage{mathtools}
\usepackage{amsmath}
\usepackage{epstopdf}
\usepackage{xspace}
\usepackage{svg}

\algnewcommand{\algorithmicgoto}{\textbf{go to}}%
\algnewcommand{\Goto}{\algorithmicgoto\xspace}%
\algnewcommand{\Label}{\State\unskip}

\title{Hierarchy of GANs for learning embodied self-awareness model}
\name{{\fontsize{9}{50}\selectfont Mahdyar Ravanbakhsh $^{1}$,  Mohamad Baydoun $^{1}$, Damian Campo $^{1}$, Pablo Marin $^{2}$, David Martin $^{2}$, Lucio Marcenaro$^{1}$, Carlo S. Regazzoni $^{1}$}}
\address{{\fontsize{10}{50}\selectfont $^{1}$ DITEN, University of Genova  \hspace{0.7 cm} $^{2}$ Carlos III University of Madrid}}
\begin{document}
%
\maketitle
\begin{abstract}
In recent years several architectures have been proposed to learn embodied agents complex self-awareness models. In this paper, dynamic incremental self-awareness (SA) models are proposed that allow experiences done by an agent to be modeled in a hierarchical fashion, starting from more simple situations to more structured ones. Each situation is learned from subsets of private agent perception data as a model capable to predict normal behaviors and detect abnormalities. Hierarchical SA models have been already proposed using low dimensional sensorial inputs. In this work, a hierarchical model is introduced by means of a cross-modal Generative Adversarial Networks (GANs) processing high dimensional visual data. Different levels of the GANs are detected in a self-supervised manner using GANs discriminators decision boundaries. Real experiments on semi-autonomous ground vehicles are presented.
\end{abstract}
\begin{keywords}
Generative adversarial networks, Multi-component models, Self-awareness modeling, Anomaly detection
\end{keywords}
\section{Introduction}
\label{sec:intro}



Self-awareness models make it possible for an agent to evaluate whether faced situations at a given time correspond to previous experiences. Self-aware computational models have been studied and several architectures have been introduced \cite{fusion18_damian,fusion18_ravan,lewis2016self,icassp2018}. Such models have to provide a framework where autonomous decisions and/or teleoperation by a human can be integrated as a capability of the device itself to dynamically evaluate the contextual situation \cite{lewis2016self}. Recent progress of signal processing and machine learning allow an agent to obtain a self-awareness model from stored multi-sensorial data coming from previously successfully completed experiences. Self-awareness layers modeling situations perceived through different sensorial modalities and can be integrated in order to build a uniform structure of cross-modal self-awareness for an agent. Using such models, the agent gains the ability to either predict the future evolution of a situation (e.g. for internal resources modulation) or to detect situations potentially unmanageable. This ``sense of the limit'' allows an agent predicting potential abnormalities with respect to the previous experiences to involve a human operator for support in due time. In this sense, the capability of detecting abnormal situations is an important feature included in self-awareness models as it can allow autonomous systems to anticipate in time their situation/contextual awareness about the effectiveness of the decision-making sub-modules \cite{Campo2017,8078511}.

In \cite{icassp2018} a two-layers self-awareness model has been proposed: Shared Layer (SL) and Private Layer (PL). The analysis of observed moving agents for understanding the normal/abnormal dynamics in a given scene from an external viewpoint is a very hot topic and emerging research field \cite{Bastani2016,Campo2017,Morris2008,Moll2010,Ermis2008,Emonet2011,Sirkin2017,icip18_damian,wacv_plug,sabokrou2016fully,rabiee2016crowd,nabi2013temporal}. One common approach is to detect abnormalities as deviations from externally observed Environment Centered (EC) models. EC models can be considered shared as they concern observation variables externally accessible to multiple agents. However, when a mobile observer agent is placed in the same EC reference system \cite{Bastani2016,Campo2017,Morris2008,Campo2017b}, the observer can also use its own private variables to detect the normal/abnormal flow of a situation under its viewpoint. The agent, however, relates its perceived private variables to actions that it performs in a given location of the environment. An abnormality can be perceived by EC models as a not-predicted EC location varying behaviour. 

However, an agent can also detect an abnormality with respect to what it is used to observe from a first-person viewpoint while performing the same task. This knowledge can be only estimated by the agent itself by means of its private perception variables. Detecting abnormalities by using a self-awareness model learned through multi-sensorial data acquired in the first-person by the agent can be possible while doing the same task for which a self-awareness SL model has been obtained. Such a model can be described as the PL of self-awareness. An external observer can have no access to such information (unless an explicit communication link is established with the agent) and will not be able to detect PL abnormalities, while he can still do this by using the SL model, if available. Thus, a well-trained model for PL self-awareness can allow an agent to be able to evaluate abnormalities with a more complete information set, based on the joint availability of PL and SL models, as it was shown in \cite{icassp2018}. However, previous works mostly rely on a high level of supervision to learn PL self-awareness models \cite{Kim2011,Bastani2016,8078511,Sirkin2017,Ramík2014,icassp2018}, while in this work, we propose a weakly-supervised method based on a hierarchy of Cross-modal Generative Adversarial Networks (GANs) for establishing self-awareness on PL. This model not only can be trained in a self-supervised manner but also can provide a level of information to boost the SL model. It can provide further normality representation for enriching the one obtained in an unsupervised way from SL normality representation and vice-versa. Both SL and PL learned models, can be used to predict the dynamics of a vehicle performing a task. They can be used also to describe normal behavioural conditions of the agent with respect to the two type of observations. Furthermore, a cross-correlation of the private and shared perspectives over the same phenomenon would provide a more complete capability for detecting the incoming anomalies \cite{Kim2011,Bastani2016}.

This paper proposes a novel method to learn the PL model using an incremental hierarchy of GANs. GANs\cite{NIPS2014_5423} are deep networks commonly used to generate data and they have shown good performance for learning the distribution of data \cite{radford2015unsupervised,DBLP:journals/corr/IsolaZZE16,icip17,ravanbakhsh2017training}. Despite the great power of GANs for modeling data distribution, it has been observed that GANs fail to learn the highly complex distributions where the data is in a high order of diversity. In other words, the GAN fails to learn the diversity, in which the generator fails to create diverse samples and the discriminator is not able to classify them as fake \cite{salimans2016improved}. This is a very known problem in the context of training GANs and several works try to tackle that \cite{srivastava2017veegan,tolstikhin2017adagan}. A natural solution to estimate data distribution with huge diversity is to break it down into smaller sets and estimate it as the mixture of multiple small distributions \cite{reynolds2015gaussian,narendra1977branch,kadir2014high}. 
The nature of an autonomous agent self-awareness model demands to learn a highly diverse distribution. The challenge in such tasks is not only the problem of learning a complex data distribution, but also the source of supervision is limited. The concept of learning a mixture of multiple distributions starting from simple to complex has been explored before \cite{abad2017autonomous,Bastani2016,sangineto2016self,abad2017self,tolstikhin2017adagan}, but they mostly rely on a high level of supervision. Inspired by \cite{tolstikhin2017adagan,sangineto2016self} we propose a self-supervised method using a set of cross-modal GANs to solve small sub-problems. This set of GANs stack in a hierarchical fashion and learns the small distributions in different levels of hierarchy. The first level is trained using the provided subset of data including most simple situations; more complex situations that can emerge in time will be abnormalities with respect to such models and their description can generate additional models. The next levels of the hierarchy are trained with the supervision provided by the first level. Namely, the scores of the first level discriminator are used to approximate the complexity of data. This can be seen as a supervision to detect whether a new model needs to be learned, and when this happens, to estimate the new model that has to be incrementally added to the hierarchy.

The main novelty of this proposed approach is a weakly-supervised strategy to divide and solve a complex problem using GANs. Furthermore, at best of our knowledge, it is the first time that the discriminator scores are used to approximate the complexity of a distribution. The discriminator scores correspond to the error (\emph{i.e.}, the innovations with respect to the other models already presented in the GANs hierarchy). As result, the method is able to model highly diverse distributions, which can be seen as a tool to find a sort of set of orthogonal basis functions by subdividing into simpler sets.

The rest of the papers is organized as follow: Sec. \ref{sec:gan_method} describes the procedure of training cross-modal GANs, and the proposed hierarchy of GANs to model the PL self-awareness. In Sec. \ref{sec:exp} results of evaluations are reported and discussed.
\begin{figure*}[t]

\begin{minipage}[b]{0.5\linewidth}
  \centering
\centerline{\includegraphics[width=\linewidth]{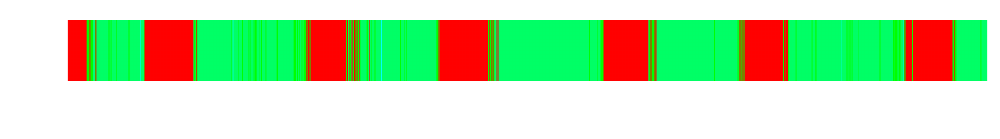}}
\vspace{-0.3cm}
  \centerline{\includegraphics[width=\linewidth]{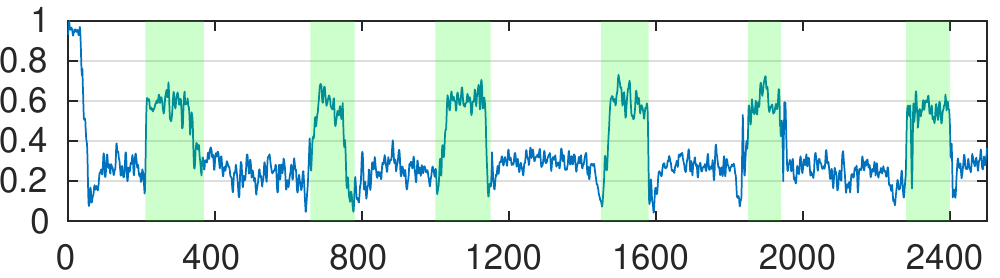}}
  \centerline{(a) Result of the first level of GANs}\medskip
\end{minipage}
\begin{minipage}[b]{0.5\linewidth}
  \centering
  \centerline{\includegraphics[width=\linewidth]{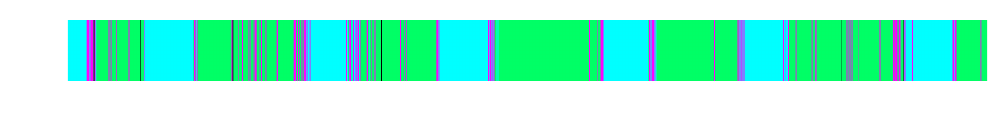}}
\vspace{-0.3cm}
  \centerline{\includegraphics[width=\linewidth]{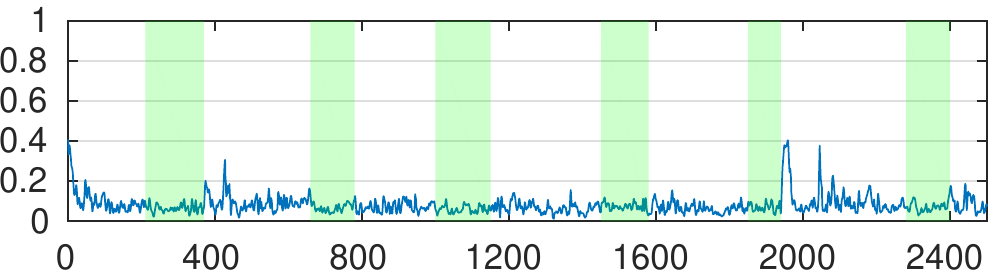}}
  \centerline{(b) Result of the hierarchy of GANs}\medskip
\end{minipage}
\caption{Training hierarchy: the horizontal axis represents the time, and the vertical axis shows the distance values. The white background means the vehicle moves straight, while the green bars indicate curving. The upper bar shows the color-coded clusters.}
\label{fig:train}
\end{figure*}
\section{AGENT EMBODIED SELF-AWARENESS MODEL}
\label{sec:gan_method}
The PL of self-awareness model consists of a hierarchical structure of cross-modal GANs. These \cite{NIPS2014_5423} are trained to learn the normality using a sequence of images synchronously collected from a first-person viewpoint paired with their corresponding optical-flow maps.
In order to understand the relation between this two modalities, the hierarchy of cross-modal GANs is adopted and trained in a weakly-supervised manner. The only supervision here is the provided subset of normal data to train the first level of the hierarchy that we called \emph{Base GAN}. The \emph{Base GAN} provides the reference for the next levels of the hierarchy and all the further levels are trained in a self-supervised manner. The rest of this section is dedicated to explaining the procedure of learning a single cross-modal GAN, constructing the hierarchy of GANs, and finally the criteria for anomaly detection using this hierarchical structure.

\noindent{\textbf{Cross-modal GANs for learning the normality}:} GANs are deep networks commonly used to generate data (e.g., images) and are trained using only unsupervised data. The supervisory information in a GAN is indirectly provided by an adversarial game between two independent networks: a generator ($G$) and a discriminator ($D$). This competition between $G$ and $D$ is helpful in boosting the ability of both $G$ and $D$. To learn the normal pattern of the observed scene two channels are used: appearance (i.e., raw-pixels) and motion (optical-flow images) for two cross-channel tasks. In the first task, optical-flow images are generated from the original frames, while in the second task appearance information is estimated from an optical-flow image.
Specifically, let $F_t$ be the $t$-th frame of a training video and $O_t$ the optical-flow obtained using $F_t$ and $F_{t+1}$. $O_t$ is computed using \cite{brox2004high}.
Two networks are trained: ${\cal N}^{F \rightarrow O}$, which is trained to generate optical-flow from frames and ${\cal N}^{O \rightarrow F}$, which generates frames from optical-flow.
In both cases, inspired by \cite{DBLP:journals/corr/IsolaZZE16,icip17}, our networks are composed by the conditional generator $G$ and the conditional discriminator $D$. $G$ takes as input an image $x$ and outputs an image of the same dimensions of $x$ but represented in a different modality.
In the proposed architecture both $G$ and $D$ are fully-convolutional networks. The $G$ network is the U-Net architecture \cite{DBLP:journals/corr/IsolaZZE16}, which is an encoder-decoder following with {\em skip connections} which help preserving important local information. 
For $D$ the {\em PatchGAN} discriminator \cite{DBLP:journals/corr/IsolaZZE16} is proposed, which is based on a ``small'' fully-convolutional discriminator $\hat{D}$. The output of $\hat{D}$ is a score map which can be seen as the encoded representation of the discriminator.
Additional details about the training can be found in \cite{DBLP:journals/corr/IsolaZZE16,icip17}.

\noindent{\textbf{Hierarchy of cross-modal GANs}:} As reviewed in Sec.\ref{sec:intro} the assumption is that the distribution of the normality patterns is under a high degree of diversity. In order to learn such distribution, we suggest a hierarchical strategy by splitting the different distributions among the different hierarchical levels, in which, each subset of train data is used to train a GAN for a different level. To construct the proposed hierarchy of GANs, a recursive procedure is adopted.
As shown in Alg. \ref{alg:hgan} the inputs of the procedure are represented by two sets: ${\cal X}$ is the entire normal sequence of training data, which includes a set of coupled Frame-Motion maps, where ${\cal X} = \{ (O_t, F_t) \}_{t=1,...,N}$, and $N$ is the number of total train samples. The input ${\cal V}_l$ is a subset of ${\cal X}$, provided to train GANs for each individual level of the hierarchy. For instance, in case of the first level GANs, the initial set ${\cal V}_0$ is used to train two cross-modal networks ${\cal N}^{F \rightarrow O}$, and ${\cal N}^{F \rightarrow O}$ (denoted by ${\cal N}$). Note that, the only supervision here is the initial ${\cal V}_0$ to train the first level of the hierarchy, and the next levels are built accordingly using the supervision provided by the first level of GANs.
After training ${\cal N}$, we input $G^{F \rightarrow O}$ and $G^{O \rightarrow F}$ (denoted by ${G}$) using each frame $F$ of the entire set ${\cal X}$ and its corresponding optical-flow image $O$, respectively. ${G} ({\cal X})$ generates the predicted Frame-Motion couples ${\cal P}$. In case of ${\cal N}^{O \rightarrow F}$, ${\cal P} = \{ (p_{O_t}, p_{F_t}) \}_{t=1,...,N}$, where $p_{O_t}$ and $p_{F_t}$ are $t$-th predicted optical-flow and predicted frame, respectively.\\
During the training phase of GANs, the output of $\hat{D}$ (the encoded representation of discriminator) over all the grid positions is averaged and this provides the final score of $D$ with respect to the input. At this time for constructing the next level of the hierarchy, we directly use the averaged scores of $\hat{D}$ as a ``detector'' which is run over the grid to detect the abnormality from the input frame. Each discriminator models the decision boundary on the learned feature space which separates the densest area of this distribution from the rest of the space. Outside this area lie both non-realistic generated images and real, unseen events. Our hypothesis is that the latter lies outside the discriminator's decision boundaries because they represent situations never observed during training and hence treated by $D$ as outliers. In other words, the distance between the discriminator encoded representation of the prediction ${\cal P}$ and the encoded layer associated with the output of $\hat{D}$ observing the new image/optical-flow couple ${\cal X}$ is higher when the input sample represents a situation never observed during training. This fact makes it possible to use the $\hat{D}$ scores as a information to detect different distributions of data to train the next levels of the GANs.

In light of the above, the distance maps ${\cal D}$ between the observations ($\cal X$) and the predictions ($\cal P$) are computed, then clustered by a self-organizing map (SOM) \cite{Kohonen2001}. Clusters with high average scores are considered as new distributions. The newly detected distributions build the new subsets ${\cal V}_{l}$ to train new GANs. All the networks ${\cal N}$ are stacked into the next levels for constructing the entire hierarchical structure ${\cal H}_{l}$. Such incremental nature of the proposed method makes it a powerful model to learn a very complex distribution of data in a self-supervised manner.\\
\begin{algorithm}
\caption{Constructing the hierarchy of GANs}\label{euclid}
\label{alg:hgan}
\begin{algorithmic}[1]
\small{
\Require
\State $\theta:  \text{  Threshold for creating a new GAN in the hierarchy}$
\State ${\cal X}:$ \text{ Entire training sequences } ${{\cal X} = \{ (O_t, F_t) \}_{t=1,...,N}}$ 
\State ${\cal V}_0: \text{  Subset of } {\cal X}$
\State $l=0: \text{  Counter of hierarchy level }$
\Ensure
\State $[{\cal H}_l] \text{ Hierarchy of GANs}$
\Procedure{CONSTRUCTING HIERARCHY OF GANS}{}\label{marker}
\Label $\texttt{train:}$
\State $\text{Train networks } {\cal N}=[{\cal N}^{F\rightarrow O}, {\cal N}^{F \rightarrow O}]\text{, with subset } {\cal V}_l$
\State $[{\cal H}_l] \gets \text{Trained networks }{\cal N} $
\State ${\cal P} \gets {G} ({\cal X}) \text{, Predictions of }\{ (p_{O_t}, p_{F_t}) \}_{t=1,...,N}$
\State ${\cal D} \gets ||\hat{D} ({\cal X}) - \hat{D} ({\cal P})||_1$
\State $\text{Cluster the distance maps: }SOM({\cal D})$

\For{$\text{each identified cluster}$}
\State $\mu \gets \text{Average score of distance maps in this cluster}$ 

\If{$\mu \ge \theta $}
\State $ l= l +1 $
\State ${\cal V}_l \gets \text{Select train samples in this cluster from }{\cal X}$
\State \Goto \texttt{train}
\EndIf
\EndFor
\State \textbf{return} $[{\cal H}_l]$
\EndProcedure
}
\end{algorithmic}
\end{algorithm}

\begin{figure}[htb]
\centerline{\scriptsize{(a) }\includegraphics[width=0.95\linewidth,height=1.5cm]{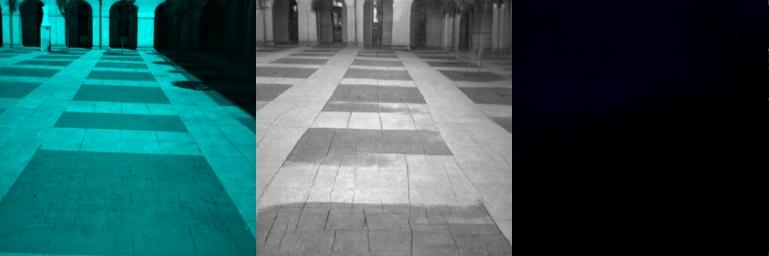}}
\centerline{\scriptsize{(b) }\includegraphics[width=0.95\linewidth,height=1.5cm]{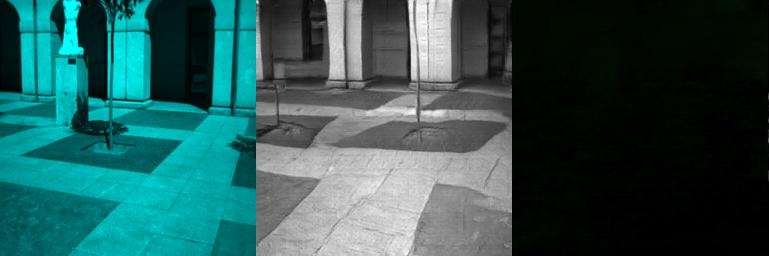}}
\centerline{\scriptsize{(c) }\includegraphics[width=0.95\linewidth,height=1.5cm]{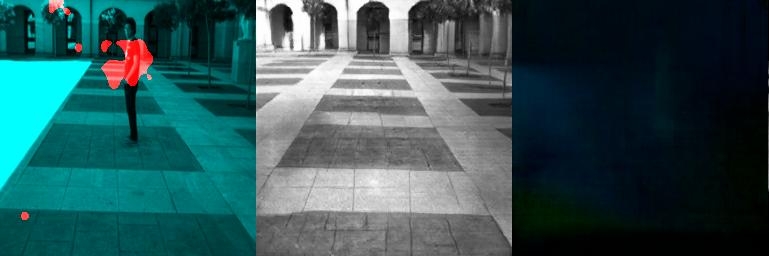}}
\centerline{\scriptsize{(d) }\includegraphics[width=0.95\linewidth,height=1.5cm]{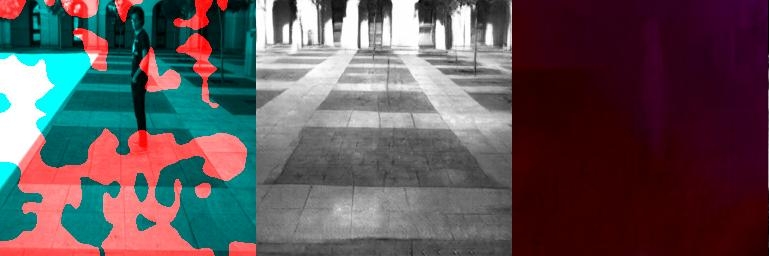}}
\centerline{\scriptsize{(e) }\includegraphics[width=0.95\linewidth,height=1.5cm]{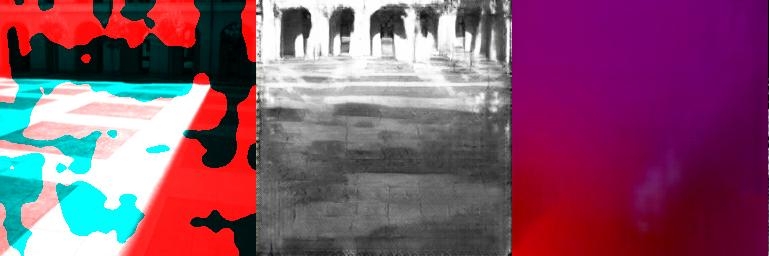}}
\caption{Anomaly visualization: the first column shows the possible abnormal areas localized on the original frame (red blobs), the second column is the predicted frame by our network, and the last column shows the pixel-by-pixel distance between the ground truth and predicted optical-flow maps. Examples of different situations are shown in (a) moving straight, (b) curving, (c) first observation of the pedestrian, and (d,e) performing the avoiding action. Note that the predictions for (c) and (d) frames have not been reconstructed the ``pedestrian'' in the corresponding reconstructions. $G^{O \rightarrow F}$ simply does not know how to ``draw'' a ``pedestrian'', since it is an unseen/abnormal situation.}
\label{fig:visavoiding}
\end{figure}
In our experiments, we use a \emph{Hierarchy of GANs} based on the identified distributions provided by $\hat{D}$ scores, which consists of two levels: $Level1$ or \emph{base GAN}, which train with a provided initialization subset ${\cal V}_0$ for representing the empty straight path, and $Level2$ which is trained over the high-scored cluster (in our case this cluster semantically representing the curves). Note that, the choice of selecting as reference the images and the optical-flow dynamics as related to situations when an agent is moving straight in an environment with no obstacle on his path is quite straightforward. In fact, different variations in the motion of the agent with respect to such a situation can be semantically associated with different interactions with the environment. By assuming that $\hat{D}$ scores deviations present a self-similarity when the same interaction happens, it can be guessed that the capability of clustering such scores into different classes can carry to an unsupervised segmentation of different interaction situations. In our formulation, the SOM’s output consists of a set of neurons encoding the main information from prediction errors and cluster them into a set of prototypes. Each of such a cluster can tell us how images and optical-flow data can be clustered depending on their similarity and on the similarity of the prediction error that a GAN produces trained on a situation when an object moves in an empty space. So the decision boundaries of the detector to describe configurations where the corrections to the generative model should be high can generate multiple regions, each characterized by a different type of self-similarities. For example, curving in an empty space can generate different prediction errors with respect to curving for avoiding an obstacle. Similarly, the discriminator's learned decision boundaries can be also used to better predict and semantically tag events at the testing time that is explained in the next section.

\noindent{\textbf{Anomaly detection}:} 
At the testing time, the discriminators are used to detect the abnormality. More specifically, input the test sample into the first level in the hierarchy of GANs, let $\hat{D}^{F \rightarrow O}$ and $\hat{D}^{O \rightarrow F}$ be the patch-based discriminators trained using the two channel-transformation tasks. 
Given a test couple ${\cal{X}}_t = {F_t, O_t}$, where $F_t$ is a test frame and its corresponding optical-flow image is $O_t$, we first produce the prediction couple ${\cal{P}}_t = {p_{O_t}, p_{F_t}}$, where ${\cal{P}}_t$ is reconstructed $p_{O_t}$ and $p_{F_t}$ using first level ${G}^{F \rightarrow O}$ and ${G}^{O \rightarrow F}$, respectively. Then, the pairs of patch-based discriminators $\hat{D}^{F \rightarrow O}$ and $\hat{D}^{O \rightarrow F}$, are applied  for the first and the second task, respectively. 
This operation results in a pair of discriminator representation for the ground truth observation: $\hat{D} ({\cal{X}}_t)$ and, the prediction: $\hat{D} ({\cal{P}}_t)$. 
Note that, a possible abnormality in the observation (e.g., an unusual object or an unusual movement) corresponds to an outlier with respect to the data distribution learned by $\hat{D}^{F \rightarrow O}$ and $\hat{D}^{O \rightarrow F}$ during training. The presence of the anomaly results in a low value of $\hat{D}^{F \rightarrow O}(p_F,p_O)$ and $\hat{D}^{O \rightarrow F}(p_O,p_F)$ (the discriminator encoded representation of prediction), but a high value of $\hat{D}^{F \rightarrow O}(F,O)$ and $\hat{D}^{O \rightarrow F}(O,F)$ (the discriminator encoded representation of observation). 
Hence, in order to decide whether an observation is normal or abnormal with respect to the scores from the current hierarchy level of GANs, we simply measure the distance between prediction and observation. The distance from the normality in the observation defines as: ${\cal{D}}_t = ||\hat{D} ({\cal{X}}_t) - \hat{D} ({\cal{P}}_t)||_1$.
\begin{figure*}[htb]
\centerline{\includegraphics[width=\linewidth]{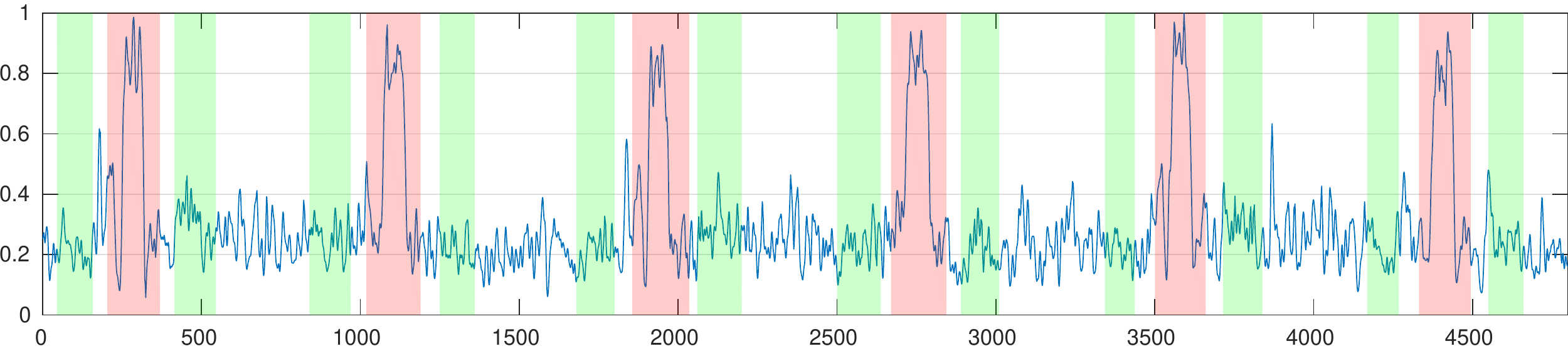}}
\caption{The abnormality signal over the test scenario. The horizontal axis represents the timeline, and the vertical axis shows the abnormality signal. The white background means that the vehicle moves on a straight line, the green bars indicate curving, and the red bars show the presence of abnormal events (a static pedestrian).}
\label{fig:testavoiding}
\end{figure*}
The computed ${\cal{D}}_t$ also determines the cluster of the test sample, by computing the distance between ${\cal{D}}_t$ and the encoded prototypes inside the SOM's neurons. Each trained neuron can be described by a centroid that encodes a known (normal) situation if ${\cal{D}}_t$ belongs to a normal cluster will be tagged as a normal sample, otherwise, it inputs to the next level of the hierarchy. The similar procedure is applied for the input sample into the next levels and eventually in the last level an error threshold is defined to detect abnormal events: when all the levels in the hierarchy of GANs tag the sample as abnormal and the measurement $\tilde{Y} = \overline{{\cal{D}}_t}$ is higher than this threshold, the current measurement is considered an abnormality. 
%
%

\section{EXPERIMENTAL RESULTS}
\label{sec:exp}
\noindent{\textbf{Dataset}:} The proposed dataset captured from an onboard camera in a real vehicle 'iCab' \cite{Marin2016}, during a perimeter monitoring task. Two different scenarios are defined: $Scenario 1$ vehicle performs a standard perimeter monitoring under the normal situation, and $Scenario 2$ the presence of anomalies while performing the perimeter monitoring task, the vehicle performs a manoeuvre to avoid a static pedestrian and continue standard patrolling. 

\noindent{\textbf{Training the hierarchy of GANs}: } The first level of GANs ($base GANs$) is trained on a selected subset of normal samples from $Scenario 1$. This subset represents the captured sequences while the vehicle moves on a straight path when the road is empty and the expected behaviour is the vehicle moving straight (normal situation). Once the pair of $base GANs$ detects an abnormality in the corresponding set on which is trained, therefore, it is expected that the corresponding observations can be considered as outliers. This is confirmed by testing the $base GANs$ over the entire sequences of $Scenario 1$, by observing the discriminators scores distances between the prediction and the observation, where ${\cal D} = ||\hat{D} ({\cal X}) - \hat{D} ({\cal P})||_1$. Fig. \ref{fig:train} shows the results of training $base GANs$. In Fig. \ref{fig:train} (a), where the test is performed using only the $base GANs$ set, it can detect the straight path (white background area) perfectly, while when the vehicle curves (green bars) it failed and recognized curving as an abnormal event. The discriminator scores distances between the prediction and the observation are higher over the curving areas, which was expected. However, after training the second level GANs using this subset of data (red cluster) and applying entire hierarchy, the model can recognize entire training sequence as normal. This happened because the different distribution of samples tagged as abnormal by the $base GANs$ input to the second level of the hierarchy, where they are recognized as normal samples.

\noindent{\textbf{Evaluations over the testing scenario}:} In order to evaluate our proposed model we apply the trained hierarchy of GANs over an unseen testing sequence ($Scenario 2$). The scenario is the moving vehicle performing the perimeter monitoring task in presence of abnormal events. In $Scenario 2$ the vehicle performs an avoidance manoeuvre over a static pedestrian and continue standard monitoring afterwards. The goal is to detect the abnormality, which is the presence of the pedestrian. In Fig. \ref{fig:testavoiding} a normalized abnormality signal is reported over the test sequence. The abnormality signal is calculated from distances between the prediction and the observation score maps. 
The red bars show the presence of an abnormal situation, which in this case is the static pedestrian. The abnormality areas start from the first sight of observing the pedestrian and continue until the avoidance manoeuvre finishes. Note that, the abnormality signal in Fig. \ref{fig:testavoiding} is computed by averaging over the distance maps: when an abnormality begins this value does not undergo large changes since the observing a compressed local abnormality (see Fig. \ref{fig:visavoiding}(c)) can not change the average value significantly. However, as soon as observing a full sight of pedestrian and starting the avoidance action by the vehicle, the abnormality signal becomes higher since both observed appearance and action are presenting an unseen situation. This situation is shown in Fig. \ref{fig:visavoiding}(d,e). 

\begin{figure}[htb]
\begin{minipage}[b]{1\linewidth}
  \centering
  \centerline{\includegraphics[width=0.98\linewidth]{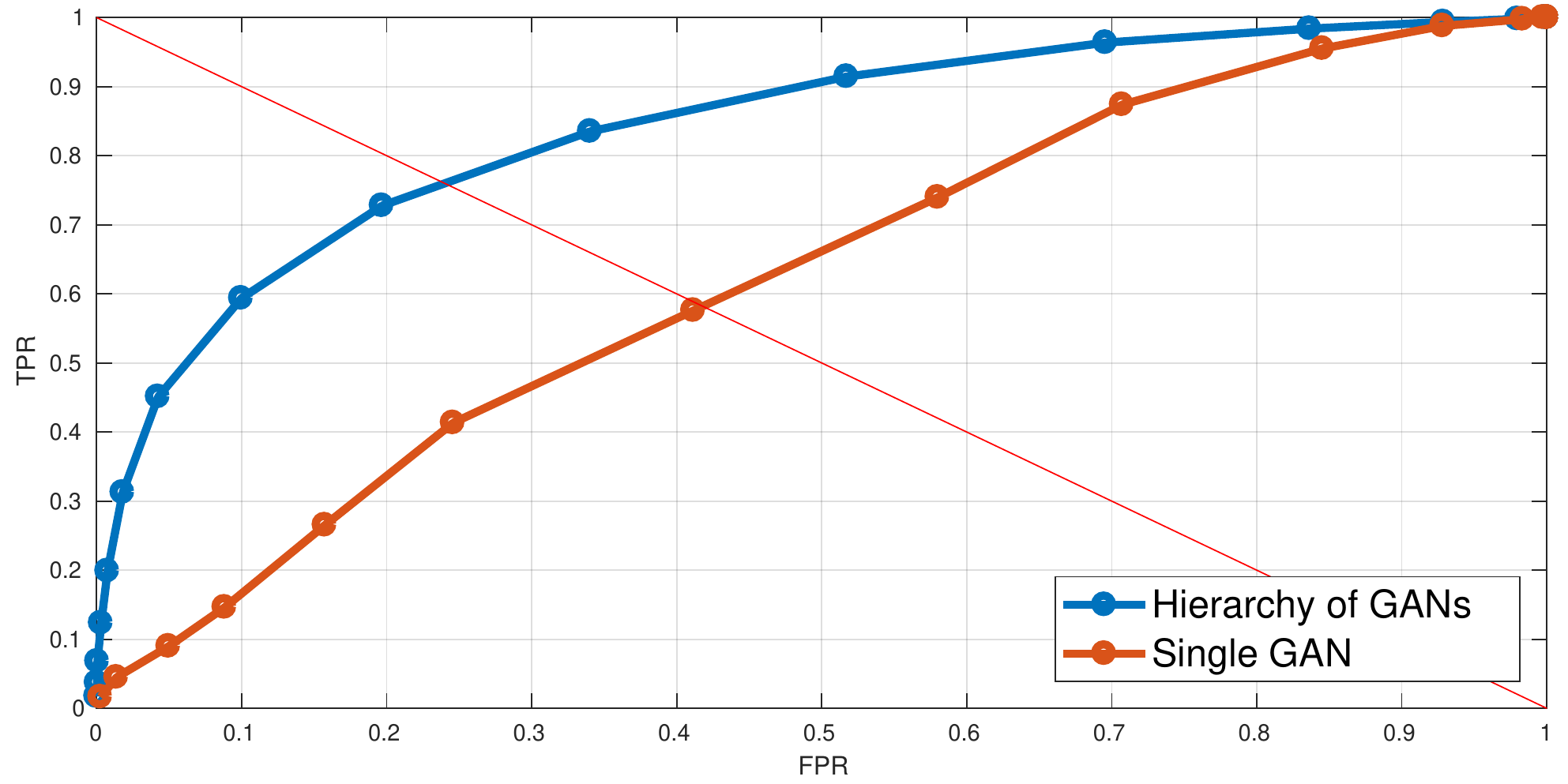}}
\end{minipage}
\caption{Single GAN vs. hierarchy of GANs for the test scenario. Comparing detection ROC curves: FPR and TPR in the horizontal axis and vertical axis, respectively.}
\label{fig:comparesingle}
\end{figure}

\noindent{\textbf{Hierarchy of GANs vs. single GAN}:} It is interesting to compare our {\em hierarchy of GANs} with the Single GAN, in other to evaluate the performance of the proposed hierarchy. In this experiments we kept all the training parameters similar to the original setup, the only difference is using entire $Scenario 1$ sequence to train the cross-modal GANs. In other words, the assumption is that a single GAN should learn entire normal pattern distribution. The result of abnormality signal is shown in Fig. \ref{fig:comparesingle} (a). However, the single GAN can detect the peak of abnormality, but the number of misdetections is higher than for the hierarchy of GANs, specifically over the curves. This could be due to the mode collapse effect which is a common issue in training GANs, the single GAN collapsed on moving straight samples and failed to generate/discriminate curving action.

Furthermore, a frame-level anomaly detection is performed for the test scenario. An abnormality label is predicted for a given test frame if at least one abnormal pixel is predicted in that frame. 
This evaluation procedure is iterated using a range of confidence thresholds in order to build a corresponding ROC curve. In our case, these confidence thresholds are directly applied to the output of the abnormality signal defined in Sec.~\ref{sec:gan_method}. Fig.~\ref{fig:comparesingle} (b) shows the ROC curves. The results using the Equal Error Rate (EER) and the Area Under Curve (AUC) for the Single GAN are: $41.90\%$ and $58.61\%$ EER and AUC, respectively, which are significantly worse than our baseline based on hierarchy of GANs: where $28.12\%$ and $79.08\%$ EER and AUC, respectively.

\section{CONCLUSIONS}
\label{sec:conc}

In this paper, a method is proposed to learning a complex data distribution based on a hierarchy of GANs in a weakly-supervised manner. This model is used to represent the PL self-awareness model of autonomous embodied agents. 
Scores of Discriminator network are used to approximate the complexity of data, which is one of our novelties. Namely, a set of distance maps between prediction and the observation scores is used as a criterion for creating another level in the hierarchical structure. Such technique facilitates breaking and solving a complex data distribution in an incremental fashion. The experimental results on semi-autonomous ground vehicles show a good performance of our method.

\bibliographystyle{IEEEbib}
\bibliography{refs}

\end{document}